\documentclass[letterpaper,twocolumn,fleqn]{article} 

\usepackage{ist}
\usepackage{graphicx}
\usepackage{float}
\usepackage{url}

\usepackage{graphicx,xcolor}
\usepackage{amsmath}
\usepackage{colortbl}
\usepackage{booktabs}
\usepackage[normalem]{ulem}
\usepackage{makecell}
\title{Color encoding in Latent Space of Stable Diffusion Models}

\author{Guillem Arias, Ariadna Solà, Martí Armengod, Maria Vanrell \\
Computer Vision Center / Universitat Autònoma de Barcelona}

\date{} 

\begin{document} 

\maketitle 

\thispagestyle{empty} 


\begin{abstract}
Recent advances in diffusion-based generative models have achieved remarkable visual fidelity, yet a detailed understanding of how specific perceptual attributes—such as color and shape—are internally represented remains limited. This work explores how color is encoded in a generative model through a systematic analysis of the latent representations in Stable Diffusion. Through controlled synthetic datasets, principal component analysis (PCA) and similarity metrics, we reveal that color information is encoded along circular, opponent axes predominantly captured in latent channels $\vec{c}_3$ and $\vec{c}_4$, whereas intensity and shape are primarily represented in channels $\vec{c}_1$ and $\vec{c}_2$. Our findings indicate that the latent space of Stable Diffusion exhibits an interpretable structure aligned with a efficient coding representation. These insights provide a foundation for future work in model understanding, editing applications, and the design of more disentangled generative frameworks.
\end{abstract}

\section{1. Introduction}
\label{sec:intro}

\noindent In the last few years, artificial intelligence (AI) has significantly progressed in building generative models able to create realistic images \cite{goodfellow2014generative,radford2016unsupervised,dhariwal2021diffusion,saharia2022imagen}. Among these, diffusion models have emerged as one of the most effective approaches, generating high-fidelity visuals by iteratively denoising random noise through learned distributions. A particularly influential architecture for this approach is Stable Diffusion \cite{rombach2022high}, which combines a latent diffusion process with a powerful image encoder-decoder pipeline.

\begin{figure*}
    \begin{center}
    \begin{tabular}{cccc}
    \includegraphics[width=0.21\linewidth]{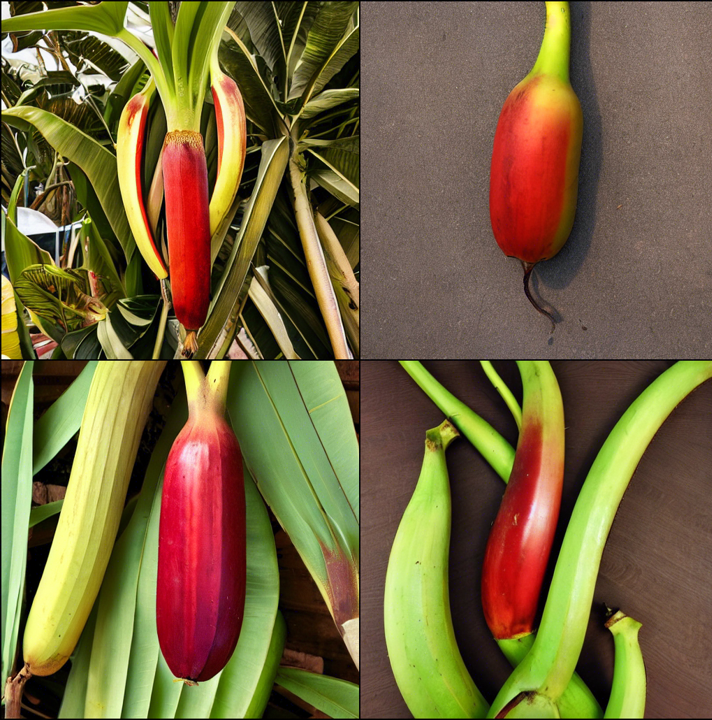} & 
    \includegraphics[width=0.21\linewidth]{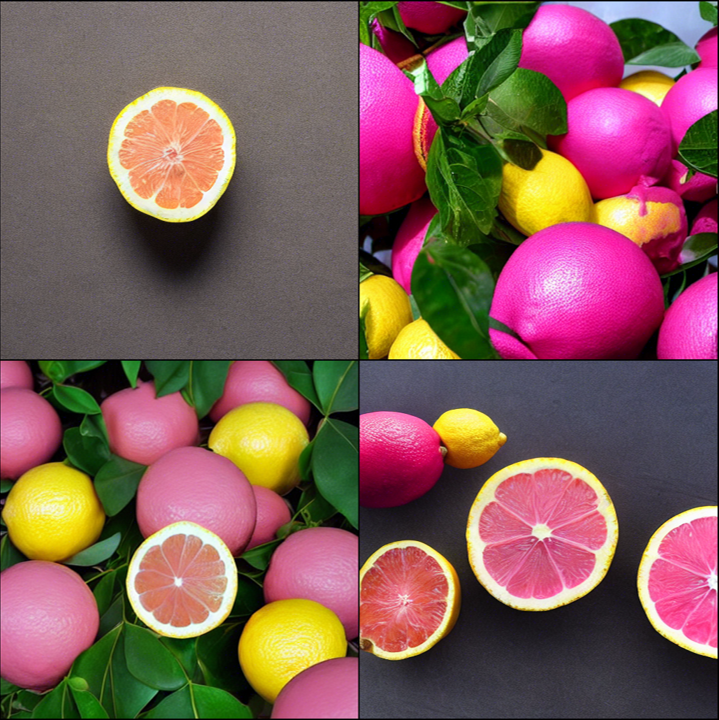} & 
    \includegraphics[width=0.21\linewidth]{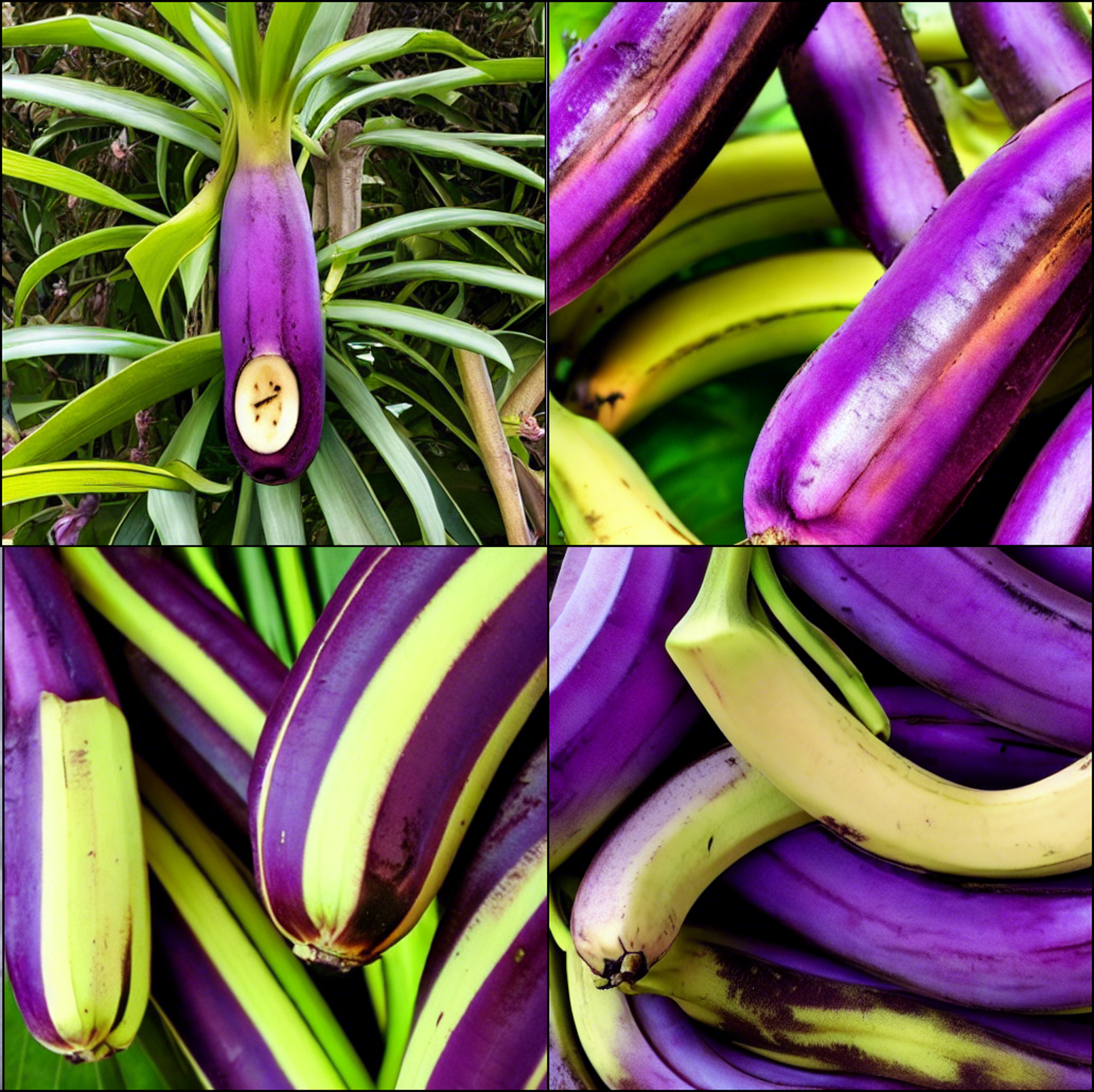} &
    \includegraphics[width=0.21\linewidth]{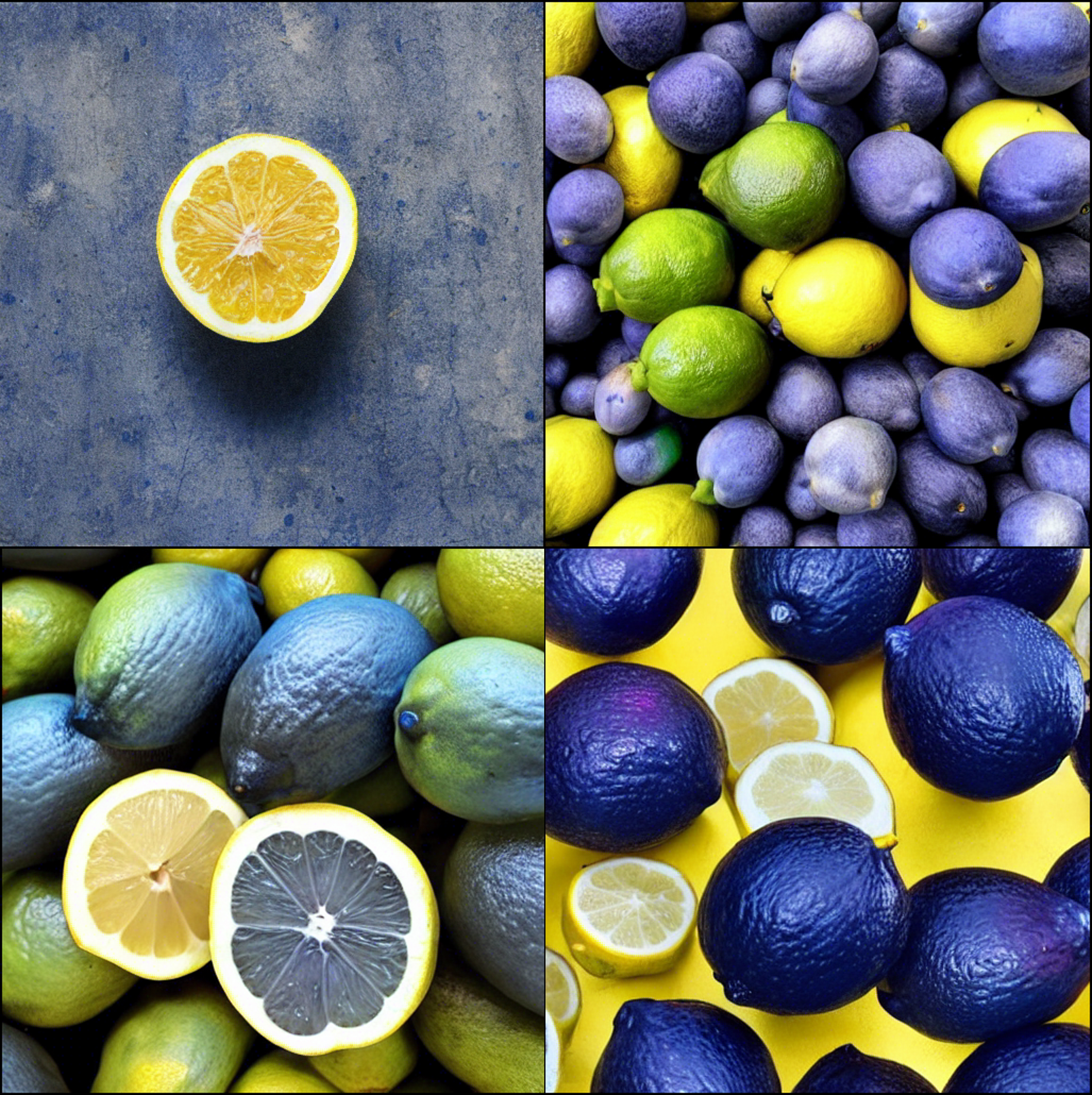} \\
    (a) \texttt{"red banana"} &
    (b) \texttt{"pink lemon"} &
    (c) \texttt{"purple banana"} &
    (d) \texttt{"purple lemon"} \end{tabular}
        \caption{Images generated by Stable-Diffusion for the given prompts formed by unnatural color-object pairs that probably were not part of the training datasets.}
    \label{fig:SDcolorobject}
    \end{center}
\end{figure*}

While much attention has been given to the quality and realism of the generated content, the mechanisms by which specific visual attributes ---such as color--- are preserved, transformed, or encoded during generation remain less explored \cite{shum2025color}. N. Moroney in \cite{CICNathan} was the first to explore different color prompts and perform some assessments. One of his conclusions is the strong preexisting color association for some objects that can provide some unexpected answers for some color prompts (see Figure \ref{fig:SDcolorobject}). The inability of the model to attribute color to any object brought us to try to understand how color and shape is encoded by these models.

This question has become increasingly relevant given recent findings that color remains one of the least controllable attributes in diffusion pipelines. Rafegas and Vanrell analyze discriminative CNNs and report opponent-like hue organization with increasing color--shape entanglement across layers \cite{rafegas2017color}, while Arias et al.\ reveal color deficiencies and text-over-vision bias in CLIP \cite{arias2025clipcolor}. 
In contrast, we study a \emph{generative} latent-diffusion pipeline and directly probe the \emph{VAE latent} that precedes diffusion, using controlled color/gray datasets, PCA, and channel ablations. 
This exposes a channel-wise mechanism—color along opponent directions mainly in $\vec{c}_3$/$\vec{c}_4$, intensity/shape in $\vec{c}_1$ (with low-frequency structure in $\vec{c}_2$)—that complements prior discriminative/VLM analyses. Recent works demonstrate that diffusion outputs often drift away from target color distributions and proposes explicit color alignment techniques to correct them \cite{shum2025color}. Similarly, new inpainting methods reveal that hue inconsistencies between filled regions and their surrounding context are a persistent failure mode \cite{wang2025inpainting}. These results highlight that reliable color handling is still not solved at the algorithmic level, motivating a deeper analysis of how color is represented at the latent level of Stable Diffusion. Moreover, color has been shown to interact with perceptual illusions in diffusion models, where emergent biases mirror phenomena observed in human vision \cite{gomezvilla2025illusions}. Understanding whether Stable Diffusion organizes chromatic information along opponent axes connects modern generative models with long-established theories of efficient coding in biological vision \cite{jammalamadaka1988correlation,pearson1895notes}.

In parallel, research has explored algorithmic approaches for controllable color editing. Methods such as content-aware color editing with auxiliary restoration tasks \cite{ren2024contentaware}, latent-guided exemplar-based recolorization \cite{yang2024latentguided}, and initialization strategies to better preserve global image properties such as background color consistency \cite{zhang2024preserveprops} have advanced recoloring capabilities. Other work focuses on inpainting directly in latent space \cite{corneanu2024latentpaint}. However, these methods treat color primarily as an external editing objective. What remains missing is a systematic account of which latent channels encode intensity versus chromatic information, and how entangled these representations are with spatial structure. This is the gap we address.

In this paper we investigate how color information is encoded in Stable Diffusion models, with a particular focus on how the image encoder projects input images into the latent space before the diffusion process begins. We aim to understand what transformations occur in the latent representations as a first step to explore color and shape encoding.

To this end, we analyze the output of the Variational Autoencoder (VAE) component used in Stable Diffusion, examining its ability to retain and reconstruct accurate color information. We also study how modifications in latent space affect color reconstruction and whether color features are encoded in disentangled or entangled representations. Previous research suggest that color attributes are tightly constrained by the realistic image distribution the model is trained on \cite{CICNathan}, making it challenging to manipulate colors as flexible, independent attributes during generation or editing.

By illuminating the mechanisms of color encoding in Stable Diffusion, this work contributes to a deeper understanding of generative AI systems and offers insights for improving their performance in applications where accurate color reproduction is essential. Our contributions on exploring the 4 channels of the latent space can be summarized as follows: 
\begin{itemize}
    \item Color presents an opponent representation based on Black-White, Cyan-Magenta, and Orange-Blue channels.
    \item Chromatic information appears to be mostly in channel 3 (Cyan-Magenta) and channel 4 (Orange-Blue), although channel 4 is strongly entangled with shape.
    \item Shape is mostly encoded in the Black-White channel 1, but channel 2 and 4 encode some shape too.
    \item In channel 4 shape is combined with orange-blue chromatic information, and channel 2 seems to be encoding more low-frequency shape information.
\end{itemize}
These results confirm again the adequacy of color-opponent representation as an efficient coding when training on natural images.

\section{2. Stable Diffusion}
\begin{figure}
    \includegraphics[width=0.9\linewidth]{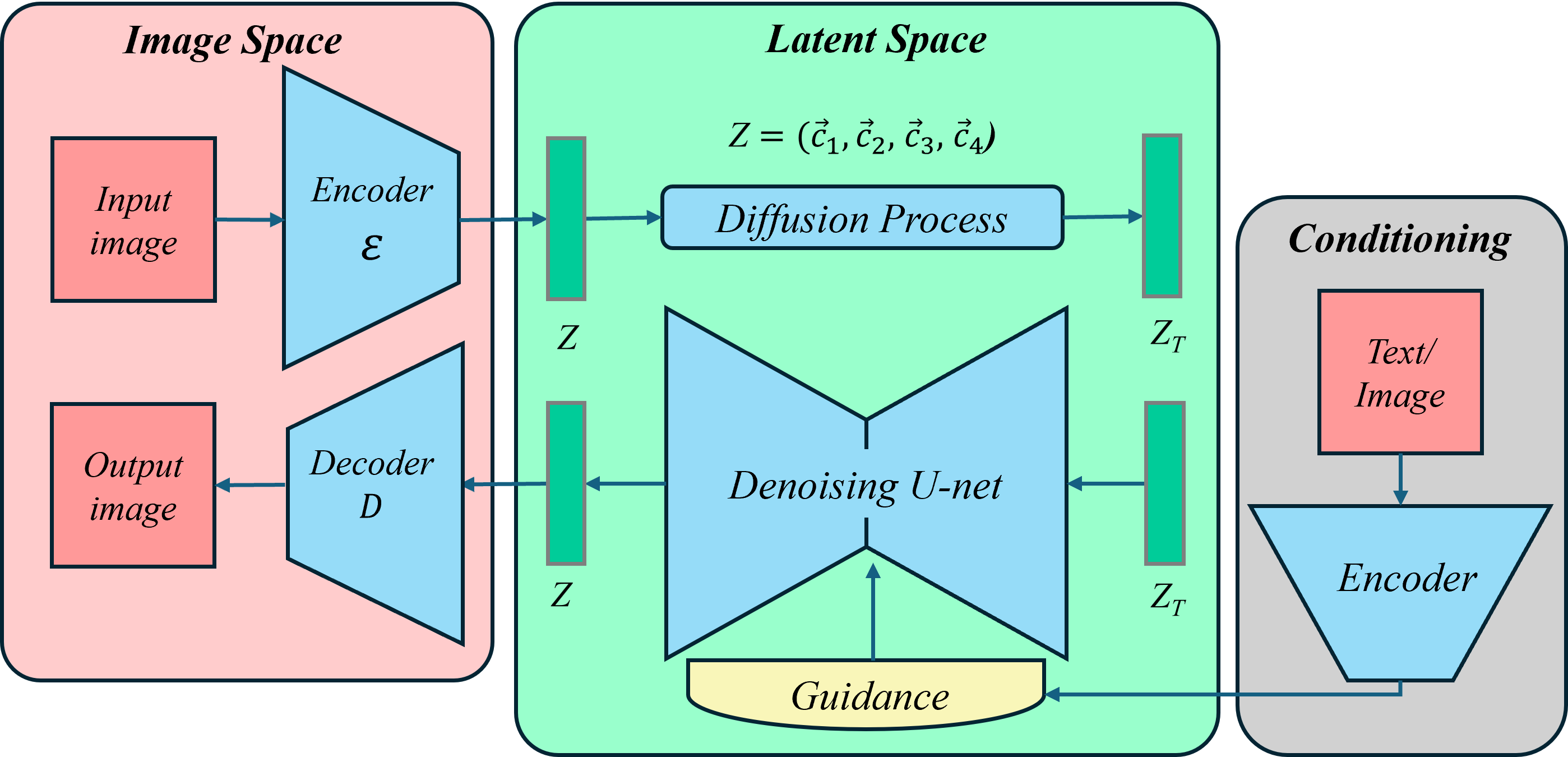}
    \caption{Stable Diffusion Model components.} \label{fig:SDmodelscheme}
\end{figure}

In this section we summarize the basic concepts of the Stable Diffusion process for image generation (see figure \ref{fig:SDmodelscheme}):

\begin{description}
    \item[Encoder:]The encoder is a CNN-based component derived from a Variational Autoencoder (VAE). It maps high-dimensional image data into a lower-dimensional latent space, compressing semantic and spatial information into a compact representation. The encoder learns a distribution over the latent space, enabling structured sampling during generation.
    \item[Latent Space:] A compact representation where spatial and semantic information is preserved, allowing efficient and scalable diffusion with reduced computational cost compared to pixel space.
    
    \item[Diffusion Process:] A forward Markov chain that incrementally adds Gaussian noise to the latent, transforming structured data into noise, thus modeling the image distribution.
    
    \item[Inverse Diffusion:] A learned denoising process (typically a U-Net) that removes noise step-by-step to reconstruct the image, predicting noise components at each timestep.
    
    \item[Conditioning and Attention:] External inputs like text or labels are injected via cross-attention layers in the U-Net, guiding generation by modulating internal representations.
    
    \item[Decoder:] A CNN-based decoder from the VAE transforms the denoised latent back into RGB image space, ensuring photorealistic reconstruction.
\end{description}

\section{3. Latent space exploration}
\noindent In contrast to previous studies that primarily focus on understanding the diffusion process itself \cite{gandikota2023concept, yang2024loracomposer,gu2023mixofshow} or the prompting process \cite{sridhar2024prompt}, here we explore how information is encoded in the latent space of diffusion models. We propose two complementary experiments to analyze this relationship. The first experiment focuses on how color is entangled in the latent representation by using a curated set of color images sampling the color space. The second experiment investigates the influence of spatial structure by using gray-level images with varying spatial properties.

\subsection{Experiment 1: Color}
\noindent To analyze how color is represented in the latent space of a diffusion model, we conducted a controlled experiment using 300 uniformly colored images spanning a wide range of hues, saturations, and intensities in the HSV color space (12 hues, 6 saturations, 5 intensities). This representation was chosen for its perceptual relevance and for its ability to separate chromatic components (hue and saturation) from luminance (value), facilitating the isolation of color-related information.

Each image was passed through the model’s image encoder, yielding a latent tensor of shape 4×64×64. Given that the input images were spatially uniform, we averaged activations across the spatial dimensions, obtaining a 4-dimensional latent vector per image. This allowed us to discard spatial variation and focus exclusively on how color is encoded in each latent channel.

We performed Principal Component Analysis (PCA) on the set of these 4-dimensional vectors to identify whether any principal components aligned with perceptual color attributes. The variance explained by each principal component is given by: \begin{equation}
(\lambda_1 ,\lambda_2, \lambda_3, \lambda_4 ) =
(0.5463, \hspace{0.2cm} 0.3172, \hspace{0.2cm} 0.1348, \hspace{0.2cm} 0.0018)
\end{equation}
where, the first three components account for over 99\% of the variance, indicating a compact, 3-dimensional encoding of color in the latent space. The eigen matrix, $M_{eig}$, is given by \begin{equation}
\begin{pmatrix}
\vec{e}_1 \\ \vec{e}_2 \\\vec{e}_3 \\ \vec{e}_4 \\
\end{pmatrix} =
\begin{pmatrix}
0.572 & 0.8058 & -0.146 & -0.0466 \\
0.1516 & -0.2781 & -0.7591 & -0.5687 \\
-0.024 & -0.0436 & -0.5917 & 0.8046 \\
-0.8058 & 0.521 & -0.2289 & -0.1641 \\
\end{pmatrix}
\end{equation} we can see that PC1 is primarily associated with channels 1 and 2, while PC2 and PC3 draw most of their contribution from channels 3 and 4.

\begin{figure}
    \centering    \includegraphics[width=0.8\linewidth]{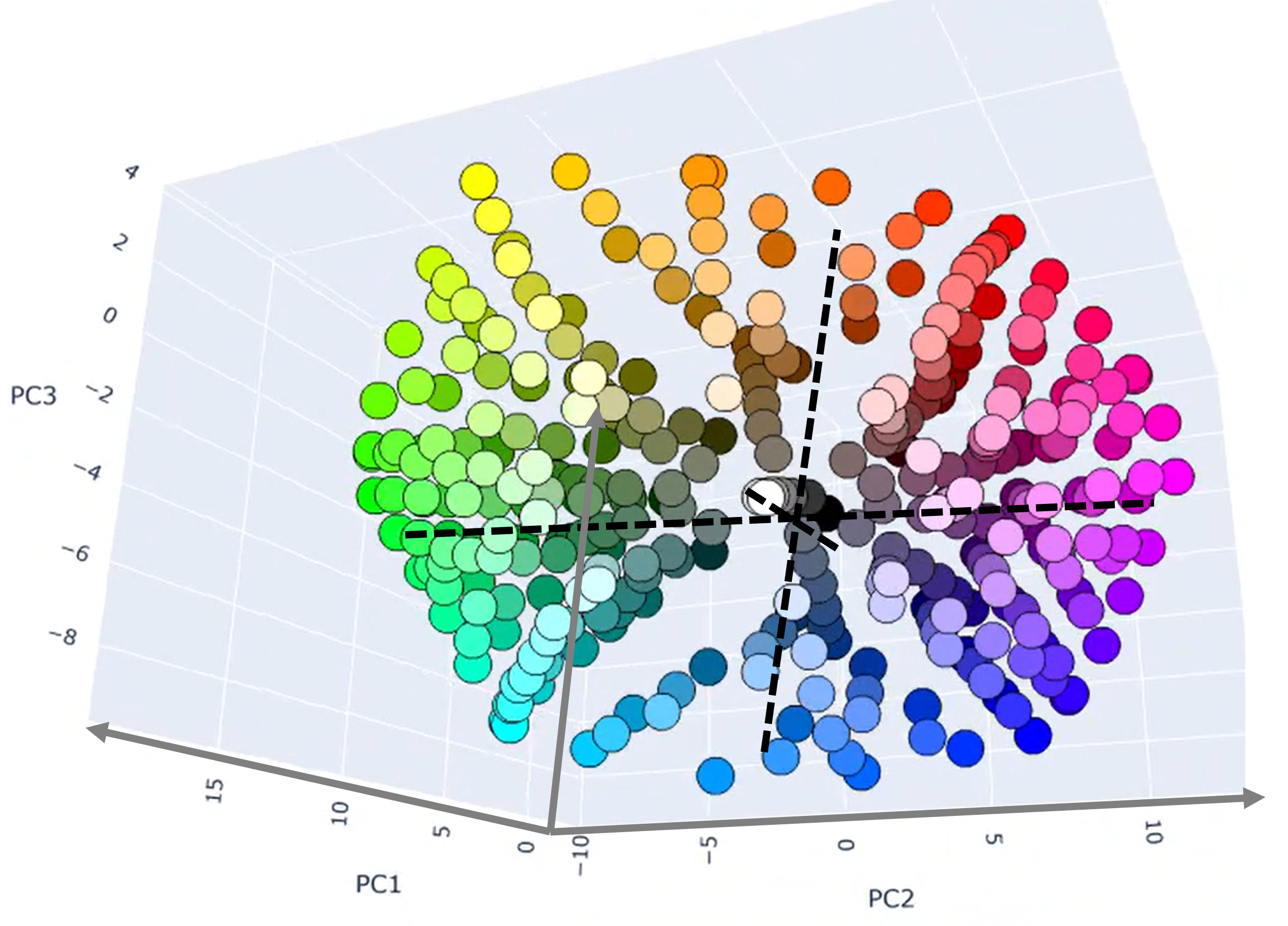}
    \caption{Homogeneous color dataset on 3 eigen vector space.} \label{fig:plot_3d}
\end{figure}

To visualize the structure of this eigen basis, we projected the latent vectors of the dataset images on the space defined by the first three principal components See figure \ref{fig:plot_3d}). The resulting embedding confirms an organized configuration, where PC1 appears to encode intensity, with the central region corresponding to achromatic images (from black to white). More notably, as it can be seen in figure \ref{fig:plot_3d} the PC2 and PC3 plane exhibits a circular structure consistent with the hue wheel, with PC2 capturing a green–magenta axis and PC3 a blue–orange one. These preliminary results indicate that color information is distributed across distinct latent channels, with intensity and chromaticity encoded along separable, interpretable directions. However, as observed in subsequent experiments, these representations may still be entangled with spatial information when spatial patterns are present.

\subsection{Experiment 2: Shape}
\noindent Following a similar analysis, we now turn to explore the representation of shape, specifically in the absence of color. Based on earlier findings suggesting that chromatic information is concentrated in specific latent channels (channels 3 and 4), we hypothesize that shape information—especially when presented in grey-level images—could be encoded primarily in the remaining channels.

To test this hypothesis, we created a dataset composed of 80 synthetic grey-scale images depicting basic geometric shapes. These shape patterns varied systematically in frequency and orientation but with no color information. Unlike in the previous color experiment, where spatial information could be averaged out, shape representation is inherently spatial. Therefore, averaging across spatial dimensions was avoided, and PCA was not applied due to the high dimensionality of the spatial space.

Instead, we assessed the contribution of each latent channel to the reconstruction of spatial structure by selectively ablating channels and measuring the degradation in image quality. Given an input image $X$, we denote its encoded latent representation\footnote{We use vectorial representation for images in  latent space.} as: \begin{equation}
\mathcal{E}(X)=\vec{Z}_X = (\vec{c}_{1,X}, \vec{c}_{2,X}, \vec{c}_{3,X}, \vec{c}_{4,X})
\end{equation} 
To estimate the importance of a specific channel (e.g., channel 3), we reconstructed the image after zeroing out that channel and computing the similarity between the original and the reconstructed image, this is \begin{equation}
\text{Similarity}(X, \mathcal{D}((\vec{0}, \vec{0}, \vec{c}_{3,X}, \vec{0})))
\end{equation}we also compute a baseline reconstruction similarity using the full latent code, \begin{equation}
\text{Similarity}(X, \mathcal{D}(\mathcal{E}(X))) = \text{Similarity}(X, \vec{Z}_X)
\end{equation}

\begin{figure} 
\begin{center}    \includegraphics[width=0.9\linewidth]{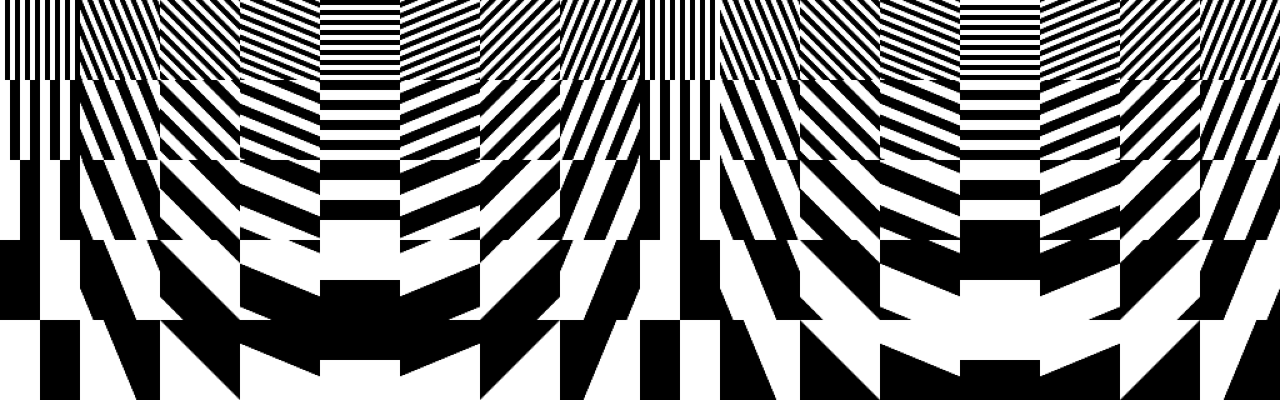}
    \caption{Image dataset used in Exp.2 to explore shape representation.} \label{fig:frequencies}
    \end{center}
\end{figure}
To quantify image similarity, we primarily use the Structural Similarity Index (SSIM) \cite{SSIM2004}, a perceptually motivated metric that correlates well with human judgment of image quality. To confirm the results, we also computed the Mean Squared Error (MSE) and Peak Signal-to-Noise Ratio (PSNR). By comparing the reconstruction quality across different combinations of retained channels (individual, pairwise, and triplet combinations), we characterized the relative importance of each latent channel for shape representation. 

\section{4. Results}
\noindent In next sections we report all the results of the above mentioned experiments. 
\subsection{Color Representation}\label{sec:resultscolor} 
\begin{figure}
\centering
\includegraphics[width=0.75\linewidth]{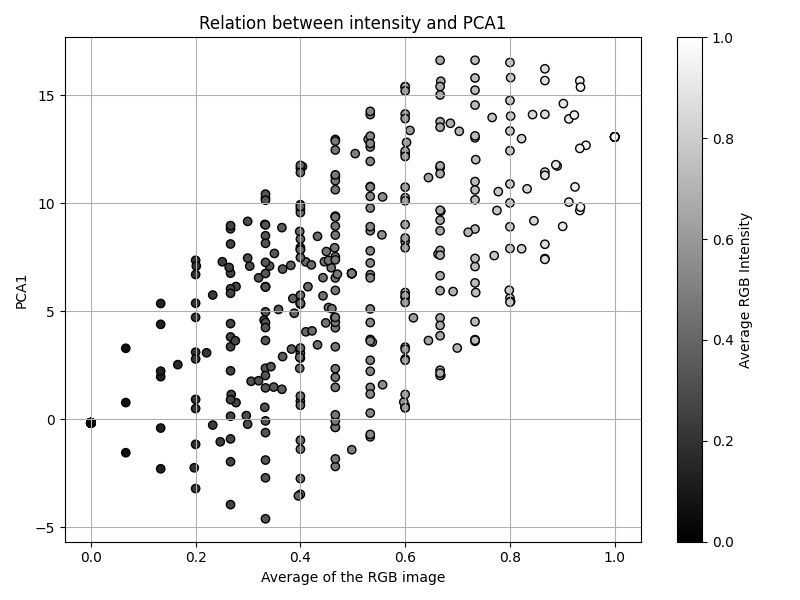}
\caption{Correlation between the first principal component (PC1) versus the mean RGB intensity of each input image.}
\label{fig:pc1_vs_HUE}
\end{figure}

\noindent In figure \ref{fig:plot_3d} we already outlined the color configuration in the latent space. To further investigate the role of the first principal component (PC1), we analyze its relationship with the average image intensity of the dataset. As shown in Figure  \ref{fig:pc1_vs_HUE}, there is a clear dependency that presents a Pearson's correlation index of 0.72\cite{pearson1895}.

Analysis of the component loadings reveals that PC1 is primarily influenced by latent channels $\vec{c_1}$ and $\vec{c_2}$, indicating that intensity-related information is predominantly encoded in these two channels.

To further investigate the representation of chromatic information in the latent space, we focused on the encoding of hue along the second and third principal components (PC2 and PC3), which primarily correspond to channels 3 and 4. To assess the distribution of hue within this subspace, we analyzed the angular relationship between PC2 and PC3 and compared it to the hue angle of the original images.

\begin{figure}
    \centering
    \includegraphics[width=0.75\linewidth]{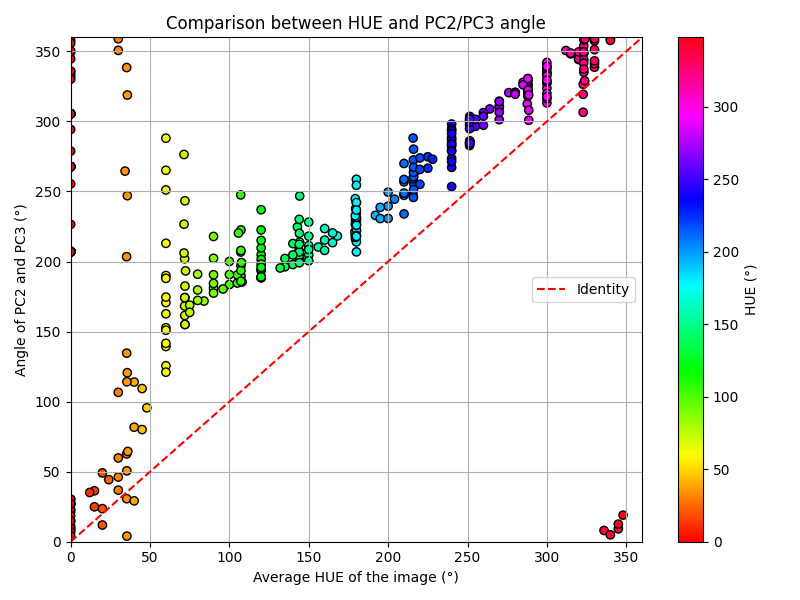}
    \caption{Correlation plot between the hue of the original images and the angle formed by the second and third principal components in latent space.}
    \label{fig:pc23_vs_HUE}
\end{figure}

As illustrated in Figure 5, the analysis reveals a close correspondence between the structure of the latent space and the hue values, indicating that hue is encoded in a directionally organized manner within the PC2-PC3 plane. This result supports the interpretation that hue is embedded in a circular latent structure, wherein variations in color are systematically aligned with angular changes in the projection space, with a circular correlations index of -0.69 (Jammalanamdak-Sarma \cite{jammalamadaka1988correlation}).

To visualize the functional role of each latent channel to both image and PCA latent space, we built a color wheel stimulus with constant intensity and varying hue. This controlled setting isolates chromatic variation and allows a targeted examination of the latent encoding. As shown in Fig. \ref{fig:pc_ch_wheel}, the first row displays decoded images obtained by projecting the original image $X$ onto each of the first four principal components. Rows 2–5 illustrate reconstructions where only a single latent channel is active, enabling an assessment of the individual contributions of $\vec{c}_1$ through $\vec{c}_4$. The bottom-left cell ($D(\vec{0})$) represents the decoding of a zero vector.

\begin{figure}
    \begin{center}      \includegraphics[width=0.9\linewidth]{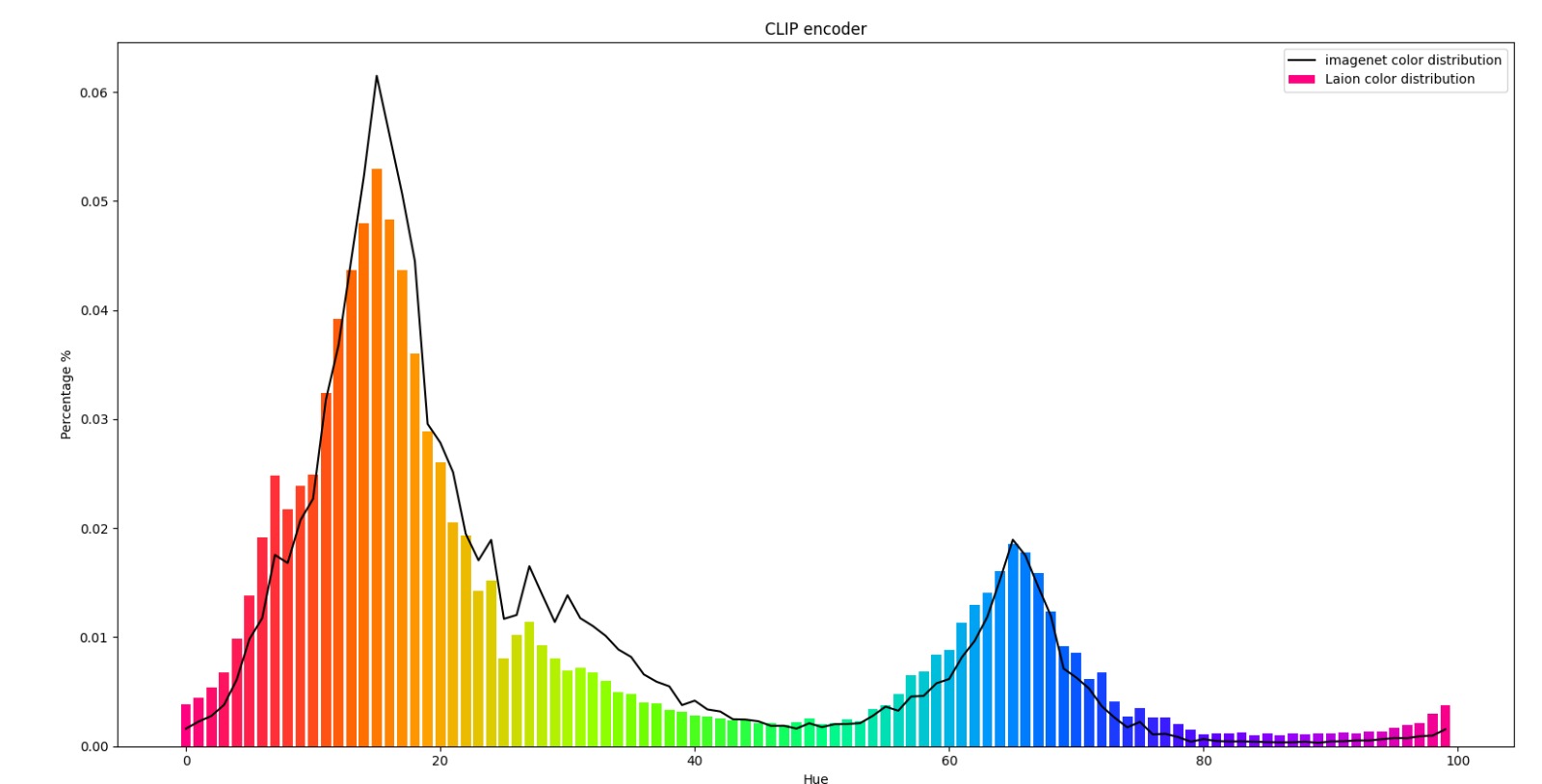}
    \caption{Hue distribution of the Dataset Laion 2B used to train Stable Diffusion model.}
    \label{fig:laion_color_distribution}
    \end{center}
\end{figure}

\begin{figure}
\centering
\small
\setlength{\tabcolsep}{1pt}
\renewcommand{\arraystretch}{0.5}
\begin{tabular}{@{\hskip 1pt}c@{\hskip 1pt}c@{\hskip 1pt}c@{\hskip 1pt}c@{\hskip 1pt}c@{\hskip 1pt}c@{\hskip 1pt}}
$X$ & $\mathcal{D}(\mathcal{E}(X))$ & $\mathcal{D}(\mathcal{E}(X)\cdot \vec{e}_1)$ & $\mathcal{D}(\mathcal{E}(X)\cdot \vec{e}_2)$ & $\mathcal{D}(\mathcal{E}(X)\cdot \vec{e}_3)$ & $\mathcal{D}(\mathcal{E}(X)\cdot \vec{e}_4)$ \\
\includegraphics[width=0.15\linewidth]{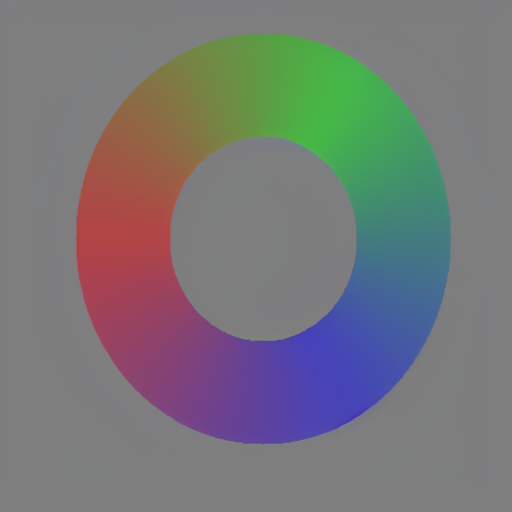} &
\includegraphics[width=0.15\linewidth]{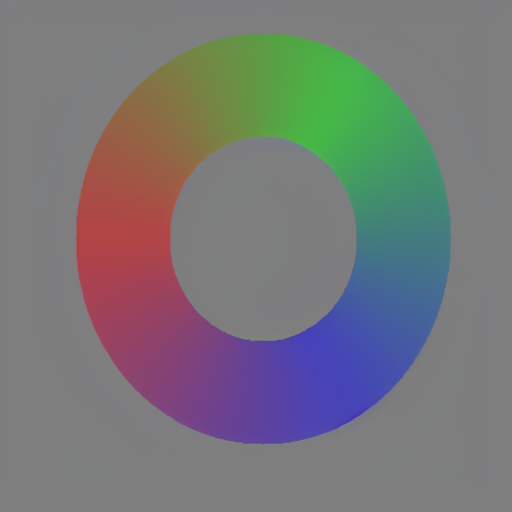} &
\includegraphics[width=0.15\linewidth]{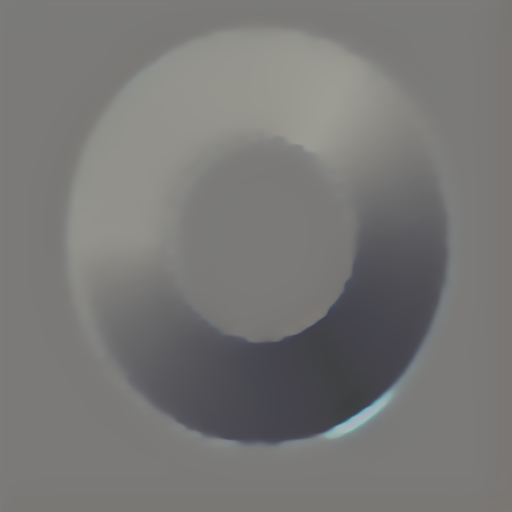} &
\includegraphics[width=0.15\linewidth]{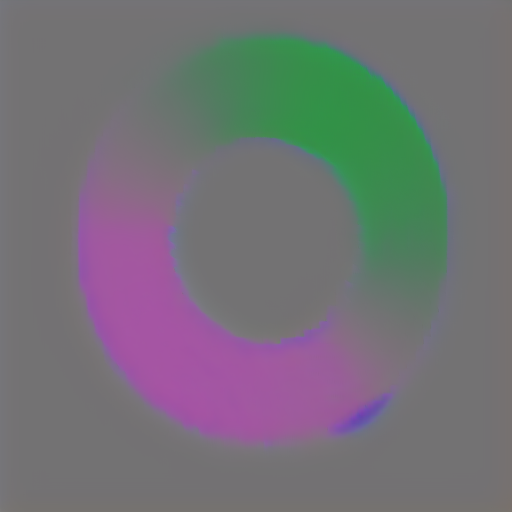} &
\includegraphics[width=0.15\linewidth]{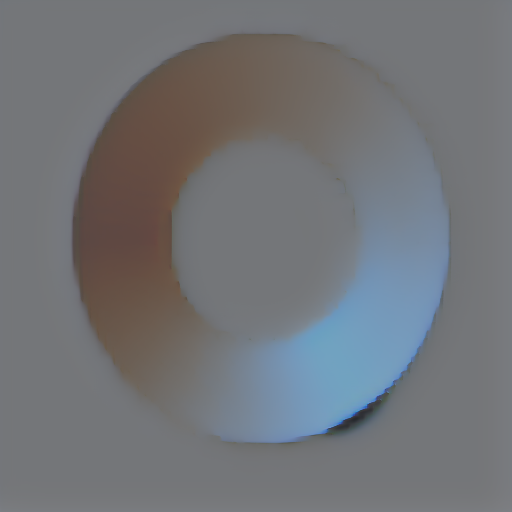} &
\includegraphics[width=0.15\linewidth]{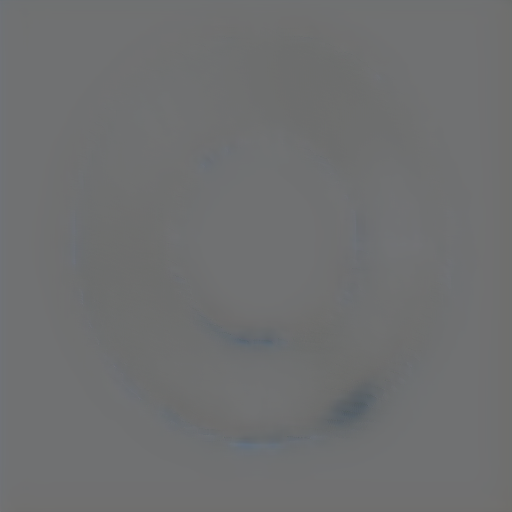} \\
\cline{2-6}
  & \includegraphics[width=0.15\linewidth]{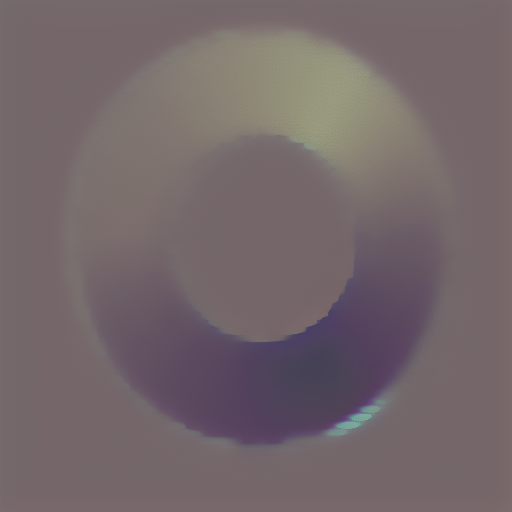} &
\includegraphics[width=0.15\linewidth]{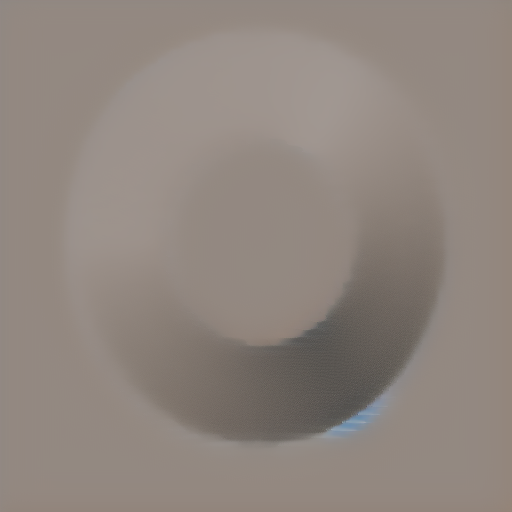} &
\includegraphics[width=0.15\linewidth]{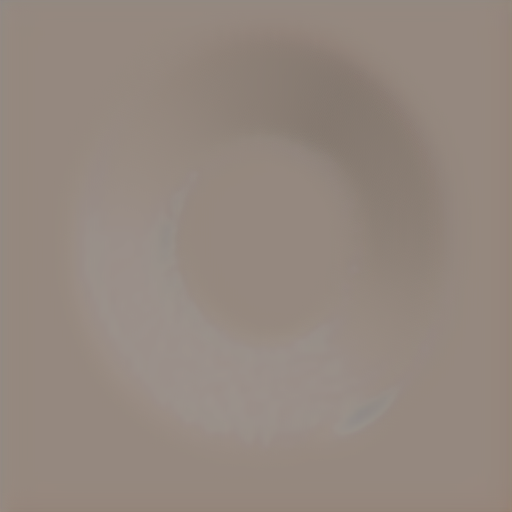} &
\includegraphics[width=0.15\linewidth]{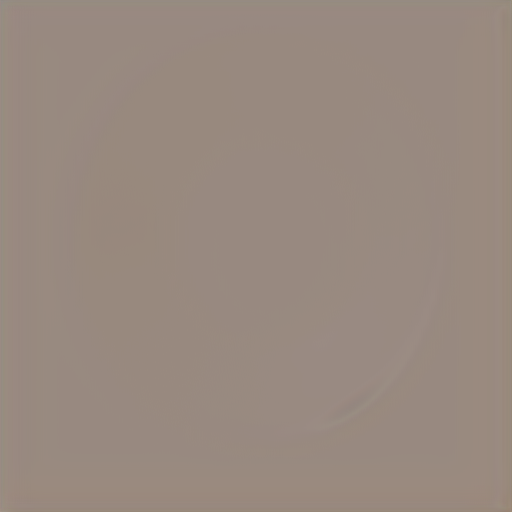} &
\includegraphics[width=0.15\linewidth]{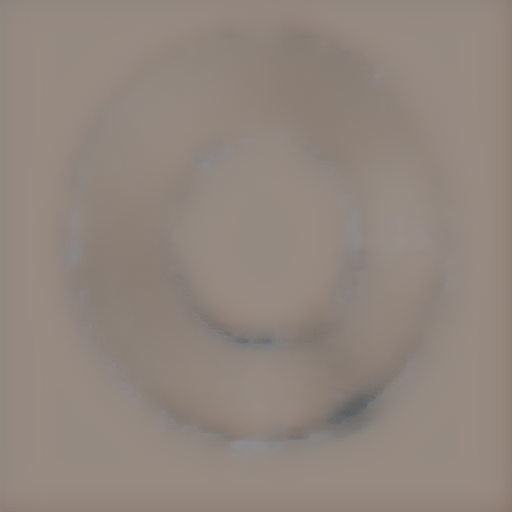} \\
 & \includegraphics[width=0.15\linewidth]{figures/FINAL/original_c2.png} &
\includegraphics[width=0.15\linewidth]{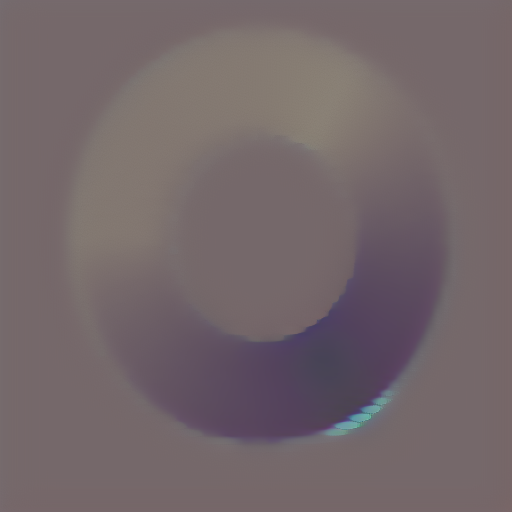} &
\includegraphics[width=0.15\linewidth]{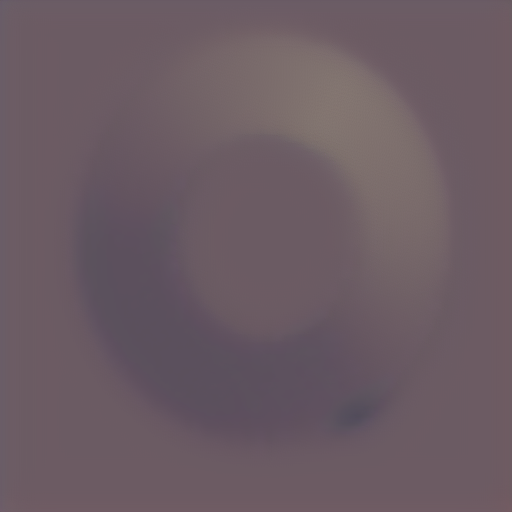} &
\includegraphics[width=0.15\linewidth]{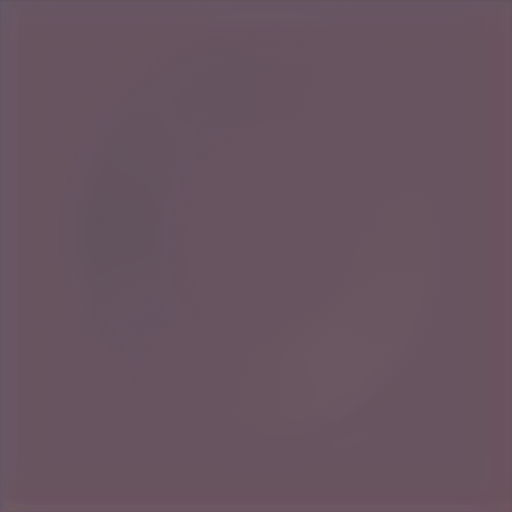} &
\includegraphics[width=0.15\linewidth]{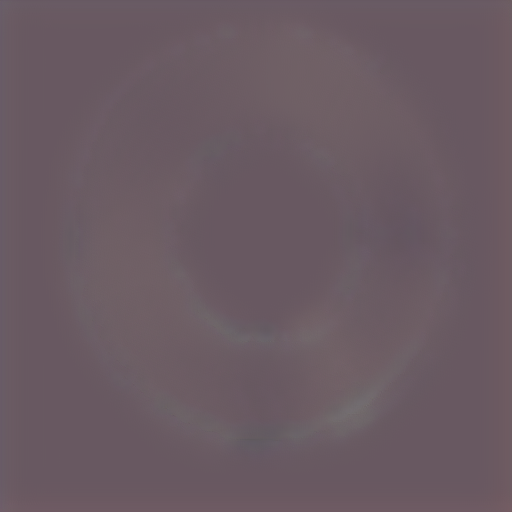} \\
$\mathcal{D}(\vec{0})$ & \includegraphics[width=0.15\linewidth]{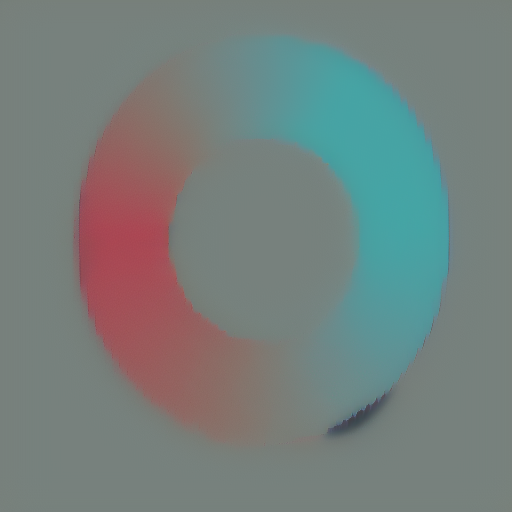} &
\includegraphics[width=0.15\linewidth]{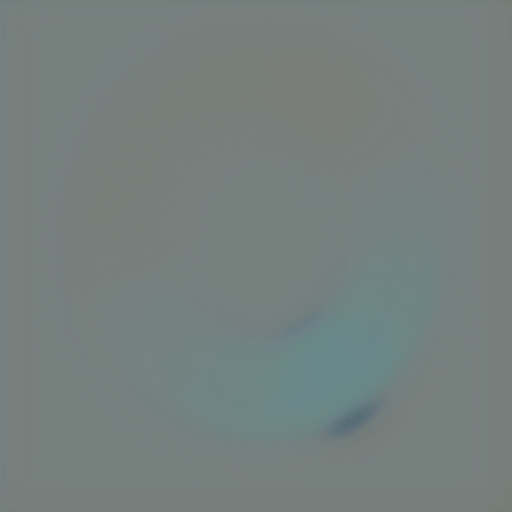} &
\includegraphics[width=0.15\linewidth]{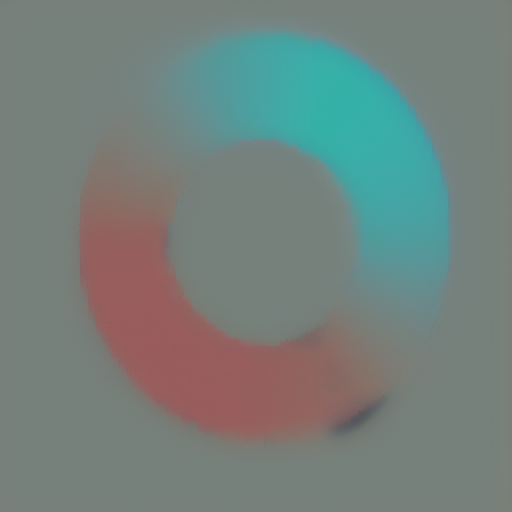} &
\includegraphics[width=0.15\linewidth]{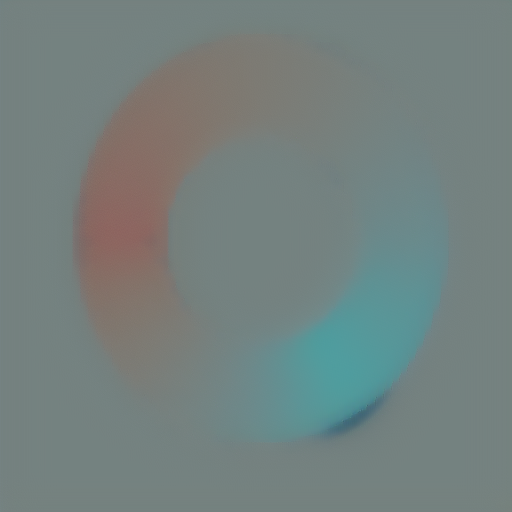} &
\includegraphics[width=0.15\linewidth]{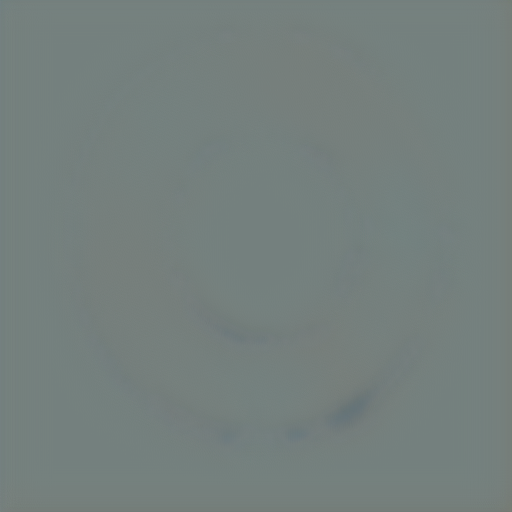} \\
\includegraphics[width=0.15\linewidth]{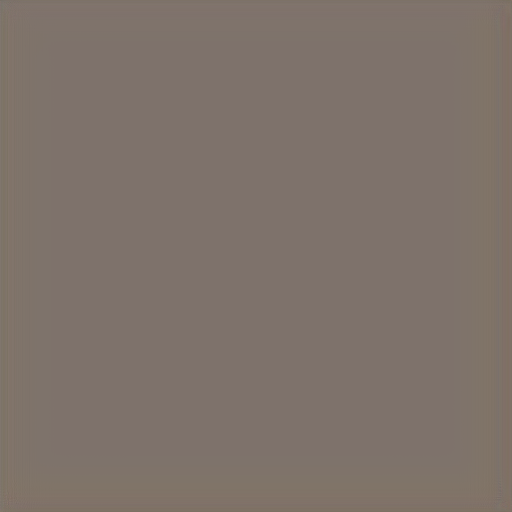} & 
\includegraphics[width=0.15\linewidth]{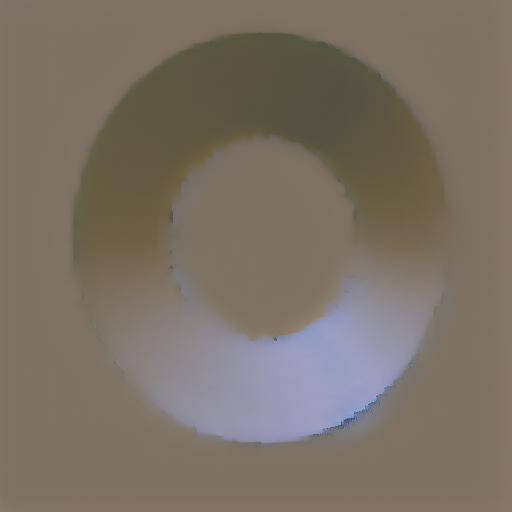} &
\includegraphics[width=0.15\linewidth]{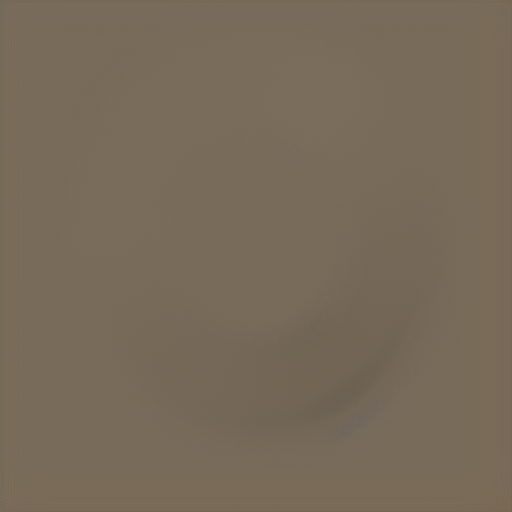} &
\includegraphics[width=0.15\linewidth]{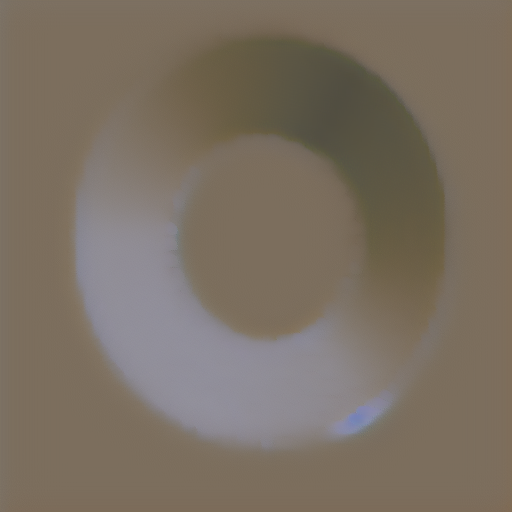} &
\includegraphics[width=0.15\linewidth]{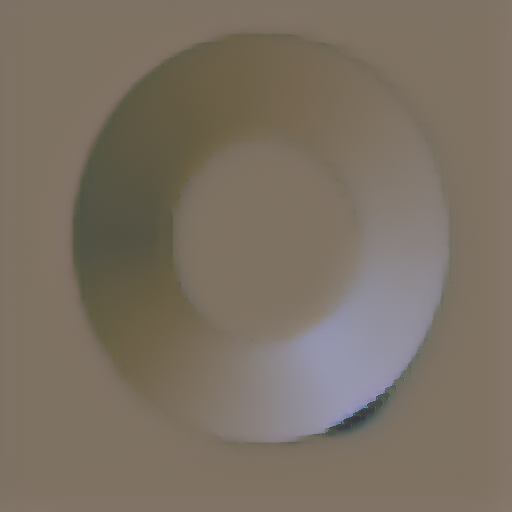} &
\includegraphics[width=0.15\linewidth]{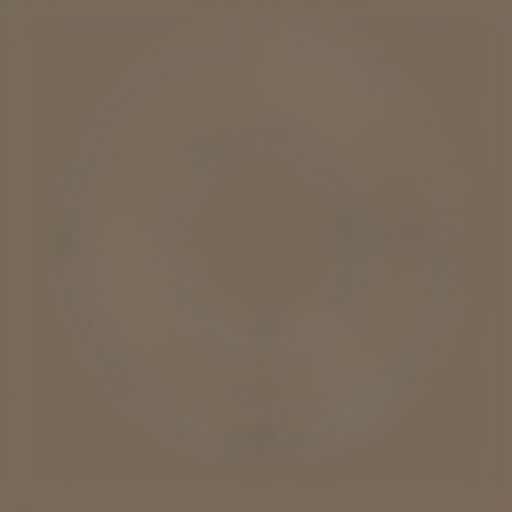} \\
\end{tabular}
\caption{Latent space color representation for the input image $X$, which varies uniformly in hue while maintaining constant intensity. The second column shows the full latent decoding $\mathcal{D}(\mathcal{E}(X))$. Columns 3 to 6 show reconstructions of $X$ projected onto each of the first four principal components. Rows 2 to 5 (except column 1) display reconstructions using only a single active latent channel. For example, row 2, column 3 shows $\mathcal{D}(\vec{c}_1, \vec{0}, \vec{0}, \vec{0})$. The bottom left image shows the decoding of a zero vector $\vec{0}$, representing the average color in the training set.}
\label{fig:pc_ch_wheel}
\end{figure}

The reconstruction from PC4 confirms its negligible contribution ($\lambda_4 = 0.018$), as no meaningful structure is observed. In contrast, PC1 captures intensity-related structure and is primarily composed of channels $\vec{c}_1$ and $\vec{c}_2$, as shown by their respective single-channel reconstructions. Channels $\vec{c}_3$ and $\vec{c}_4$ do not contribute to PC1, confirming their lack of involvement in intensity encoding. PC2 reveals an opponent axis corresponding to Magenta–Green, predominantly encoded in channel $\vec{c}_3$. PC3 encodes a second opponent axis, Blue-Orange, primarily represented in channel $\vec{c}_4$, with some contribution from $\vec{c}_3$. These axis would correlate with the peaks of the color distribution of the dataset where the model was trained on, which is given in Figure\ref{fig:laion_color_distribution}.

Overall, the analysis highlights a latent space where intensity is mainly encoded in channels $\vec{c}_1$ and $\vec{c}_2$, and chromaticity is captured across channels $\vec{c}_3$ and $\vec{c}_4$, reflecting a circular, opponent-based chromatic representation.

\subsection{Shape Representation}
\noindent In Table~\ref{tab:imatges-simples} we present the results of our similarity metrics calculated on the grey-scale shape dataset. The left side of the table displays SSIM, PSNR, and MSE values for all possible combinations of active latent channels, along with the percentage of structural information recovered relative to the minimum and maximum cases. The right side of the table organizes the same results based on the number of active channels per combination.

\begin{table}
\centering
\begin{tabular}{|l|c|c|c|}
\toprule
\textbf{Channels} & \textbf{SSIM (\%)} & \textbf{PSNR} & \textbf{MSE$\downarrow$} \\
\midrule
$[\vec{0},\vec{0},\vec{0},\vec{0}]$ & \textbf{0.3522 (0\%)} & 6.05 & 0.2481 \\
$[\vec{c}_{1},\vec{c}_{2},\vec{c}_{3},\vec{c}_{4}]$ & \textbf{0.4855 (100\%)} & 11.42 & 0.0721\\ \hline
$[\vec{0},\vec{0},\vec{c}_{3},\vec{0}]$ & 0.3528 (4.34\%) & 12.70 & 0.0617 \\
$[\vec{0},\vec{c}_{2},\vec{0},\vec{0}]$ & 0.3581 (4.43\%) & 8.74 & 0.1338 \\
$[\vec{0},\vec{0},\vec{0},\vec{c}_{4}]$ & 0.4263 (55.59\%) & 9.10 & 0.1232 \\
$[\vec{c}_{1},\vec{0},\vec{0},\vec{0}]$ & 0.4478 (71.72\%) & 9.24 & 0.1192 \\ \hline
$[\vec{0},\vec{c}_{2},\vec{c}_{3},\vec{0}]$ & 0.3621 (7.43\%) & 8.49 & 0.1417 \\
$[\vec{0},\vec{0},\vec{c}_{3},\vec{c}_{4}]$ & 0.4285 (57.24\%) & 8.95 & 0.1273 \\
$[\vec{c}_{1},\vec{0},\vec{c}_{3},\vec{0}]$ & 0.4427 (67.89\%) & 8.50 & 0.1413 \\
$[\vec{0},\vec{c}_{2},\vec{0},\vec{c}_{4}]$ & 0.4437 (68.64\%) & 10.66 & 0.0860 \\
$[\vec{c}_{1},\vec{0},\vec{0},\vec{c}_{4}]$ & 0.4716 (89.57\%) & 10.78 & 0.0835 \\
$[\vec{c}_{1},\vec{c}_{2},\vec{0},\vec{0}]$ & 0.4721 (89.95\%) & 10.64 & 0.0864 \\ \hline
$[\vec{0},\vec{c}_{2},\vec{c}_{3},\vec{c}_{4}]$ & 0.4469 (71.04\%) & 10.71 & 0.0850 \\
$[\vec{c}_{1},\vec{c}_{2},\vec{c}_{3},\vec{0}]$ & 0.4724 (90.17\%) & 10.44 & 0.0904 \\
$[\vec{c}_{1},\vec{0},\vec{c}_{3},\vec{c}_{4}]$ & 0.4729 (90.55\%) & 10.68 & 0.0856 \\
$[\vec{c}_{1},\vec{c}_{2},\vec{0},\vec{c}_{4}]$ & 0.4827 (97.9\%) & 11.40 & 0.0725 \\
\bottomrule
\end{tabular}
\caption{Table 1. Similarity metrics between original images and their reconstructions using all combinations of latent channels, sorted by SSIM.}
\label{tab:imatges-simples}
\end{table}

To define the reference points for recovery percentages, we treat the decoded output using all four latent channels as the maximum reference (100\% SSIM recovery), and the output with all channels zeroed out as the minimum (0\%). As expected, the complete reconstruction (all channels active) achieves the highest SSIM value (0.4855), while the zero-latent input yields the lowest (0.3522).

We begin by analyzing the individual contributions of each latent channel. Channel $\vec{c}_1$ is the most informative when used alone, recovering 71.72\% of the total possible SSIM. This supports our hypothesis that structural (shape) information is strongly encoded in this channel. In contrast, channel $\vec{c}_3$, which was previously associated with color encoding, contributes the least to the reconstruction of grey-scale images. This is consistent with the findings in Section~\ref{sec:resultscolor}. Channel $\vec{c}_2$ performs only slightly better, suggesting limited involvement in shape encoding. Interestingly, channel $\vec{c}_4$ recovers over 55\% of the structural information despite its established role in color representation, suggesting a dual encoding role.

When two channels are combined, the combination of $\vec{c}_3$ and  $\vec{c}_2$ fall below 10\%. However, the rest of pairs when  $\vec{c}_1$ or  $\vec{c}_4$ are present, the percentages grow significantly. Nonetheless, it is not a linear growing, since $[\vec{c}_1,\vec{c}_2,\vec{0},\vec{0}]$ achieves the maximum 89.95\% SSIM recovery over the combination of $\vec{c}_1$ and $\vec{c}_4$.

Three-channel combinations reveal similar trends. Combinations lacking $\vec{c}_1$ show a significant reduction of similarity. For instance, $[\vec{0},\vec{c}_2,\vec{c}_3,\vec{c}_4]$ recovers only 71.04\%, whereas $[\vec{c}_1,\vec{c}_2,\vec{0},\vec{c}_4]$ achieves 97.9\%. These results consistently point to $\vec{c}_1$ as the principal carrier of shape-related information in the latent space.
\begin{table}
\begin{center}
\begin{tabular}{|c|c||c|c|}
\hline
\multicolumn{2}{|c||}{\textbf{Low Frequency}} & \multicolumn{2}{c|}{\textbf{High Frequency}} \\
\hline
\textbf{\small{Channels}} & \textbf{SSIM (\%)}  &
\textbf{\small{Channels}} & \textbf{SSIM (\%)}  \\
\hline
$[\vec{0}, \vec{0}, \vec{0}, \vec{0}]$ & 0.3669 (0.00) & \cellcolor{yellow!30}$[\vec{0}, \vec{c_2}, \vec{0}, \vec{0}]$ & \cellcolor{yellow!30}0.2506  (0.00) \\
$[\vec{0}, \vec{0}, \vec{c_3}, \vec{0}]$ & 0.3702 (2.92) & \cellcolor{yellow!30}$[\vec{0}, \vec{c_2}, \vec{c_3}, \vec{0}]$ & \cellcolor{yellow!30}0.2605 (3.86) \\
\cellcolor{yellow!30}$[\vec{0}, \vec{c_2}, \vec{c_3}, \vec{0}]$ & \cellcolor{yellow!30}0.3985 (27.96) & $[\vec{0}, \vec{0}, \vec{c_3}, \vec{0}]$ & 0.2898 (15.28) \\
\cellcolor{yellow!30}$[\vec{0}, \vec{c_2}, \vec{0}, \vec{0}]$ & \cellcolor{yellow!30}0.3991 (28.49) & $[\vec{0}, \vec{0}, \vec{0}, \vec{0}]$ & 0.2975 (18.28) \\
$[\vec{0}, \vec{0}, \vec{c_3}, \vec{c_4}]$ & 0.4396 (64.34) & $[\vec{0}, \vec{0}, \vec{0}, \vec{c_4}]$ & 0.3717 (47.21) \\
$[\vec{0}, \vec{0}, \vec{0}, \vec{c_4}]$ & 0.4396 (64.34) & $[\vec{0}, \vec{0}, \vec{c_3}, \vec{c_4}]$ & 0.3856 (52.63) \\
$[\vec{c_1}, \vec{0}, \vec{c_3}, \vec{0}]$ & 0.4514 (74.78) & $[\vec{c_1}, \vec{0}, \vec{c_3}, \vec{0}]$ & 0.4095 (61.95) \\
\cellcolor{yellow!30}$[\vec{0}, \vec{c_2}, \vec{0}, \vec{c_4}]$ & \cellcolor{yellow!30}0.4556 (78.50) & $[\vec{c_1}, \vec{0}, \vec{0}, \vec{0}]$ & 0.4098 (62.07) \\
$[\vec{c_1}, \vec{0}, \vec{0}, \vec{0}]$ & 0.4569 (79.65) & \cellcolor{yellow!30}$[\vec{0}, \vec{c_2}, \vec{0}, \vec{c_4}]$ & \cellcolor{yellow!30}0.4129 (63.27) \\
\rowcolor{yellow!30}$[\vec{0}, \vec{c_2}, \vec{c_3}, \vec{c_4}]$ & 0.4574 (80.09) & $[\vec{0}, \vec{c_2}, \vec{c_3}, \vec{c_4}]$ & 0.4225 (67.02) \\
$[\vec{c_1}, \vec{c_2}, \vec{c_3}, \vec{0}]$ & 0.4702 (91.42) & $[\vec{c_1}, \vec{0}, \vec{0}, \vec{c_4}]$ & 0.4645 (83.39) \\
$[\vec{c_1}, \vec{c_2}, \vec{0}, \vec{0}]$ & 0.4705 (91.68) & $[\vec{c_1}, \vec{0}, \vec{c_3}, \vec{c_4}]$ & 0.4726 (86.55) \\
$[\vec{c_1}, \vec{0}, \vec{c_3}, \vec{c_4}]$ & 0.4729 (93.81) & $[\vec{c_1}, \vec{c_2}, \vec{0}, \vec{0}]$ & 0.4770 (88.27) \\
$[\vec{c_1}, \vec{0}, \vec{0}, \vec{c_4}]$ & 0.4729 (93.81) & $[\vec{c_1}, \vec{c_2}, \vec{c_3}, \vec{0}]$ & 0.4804 (89.59) \\
$[\vec{c_1}, \vec{c_2}, \vec{0}, \vec{c_4}]$ & 0.4785 (98.76) & $[\vec{c_1}, \vec{c_2}, \vec{0}, \vec{c_4}]$ & 0.4983 (96.57) \\
\small{$[\vec{c_1}, \vec{c_2}, \vec{c_3}, \vec{c_4}]$} & 0.4799 (100.0) & \small{$[\vec{c_1}, \vec{c_2}, \vec{c_3}, \vec{c_4}]$} & 0.5071 (100.0) \\
\hline
\end{tabular}
\caption{Table 2. Comparison of SSIM and relative improvement (\%) for different channel combinations, separated by low and high frequency images.}
\label{tab:ssim-results}
\end{center}
\end{table}

Previous results, make us to conclude that channel $\vec{c}_2$ is the more unexplained. We try to further analyze it considering its behaviour under different frequency conditions. Table~\ref{tab:ssim-results} show the results of analysising for low- and high-frequency images separately. In low-frequency cases, activations involving $\vec{c}_2$ contribute notably to structural recovery---up to 28.5\% improvement when paired with $\vec{c}_3$. However, in high-frequency cases, the same combinations show minimal contribution (0--3.8\%). This suggests that $\vec{c}_2$ may play a more active role in encoding low-frequency shape information, that would be more global shape information than image details.

Moreover, combinations of $\vec{c}_2$ with $\vec{c}_4$ are particularly effective in low-frequency settings, recovering 78--80\% SSIM, but are less impactful in high-frequency contexts (63--67\%). Finally, combinations including both $\vec{c}_1$ and $\vec{c}_2$ consistently produce high recovery scores in both frequency bands, exceeding 88\% in all cases. These findings suggest that while shape information is concentrated in $\vec{c}_1$, there can be some structures that can be distributed across other channels, particularly $\vec{c}_2$ and $\vec{c}_4$.

In summary, the results confirm a partial specialization of latent channels: $\vec{c}_1$ is primarily responsible for encoding shape; $\vec{c}_3$ is mostly color-related, but with some color information in $\vec{c}_4$. This last one, jointly with $\vec{c}_2$ appear to play more nuanced roles in shape representation aspects.

\section{Conclusions}

This work presents a detailed analysis of how color and shape are encoded in the latent space of Stable Diffusion models. Through controlled experiments on the latent representations of specific datasets. We extract the following key findings. Color presents an opponent representation in the latent space based on 3 main channels: Black-White, Magenta-Green and Blue-Orange. The 4 channels of the latent space exhibit a partially disentangled representation that can briefly be summarized as: $\vec{c}_1$ carries the bulk of intensity and shape information, $\vec{c}_2$ contributes to low-frequency structures and complements $\vec{c}_1$ with shape, $\vec{c}_3$ is a pure chromatic channel, and $\vec{c}_4$, while chromatic in nature, also exhibit varying degrees of spatial entanglement. 

With this work we are just giving some preliminary steps towards the understanding of color and shape representation, further research is required to condition their generation. 
The results presented here demonstrate the feasibility of precise control over visual attributes such as hue, saturation, and shape by selectively manipulating latent channels. They also point to the limitations in achieving fully disentangled representations using standard training procedures on natural image data.

\section{Acknowledgments}
This work was partially supported by grants PID2021-128178OB-I00, PID2024-162555OB-I00 funded by MCIN/AEI/10.13039/501100011033 and by ERDF/EU, and by the Generalitat de Catalunya --- Departament de Recerca i Universitats with reference 2021SGR01499 and CERCA Program.

\begin{figure}[H]
\centering
\includegraphics[width=0.6\linewidth]{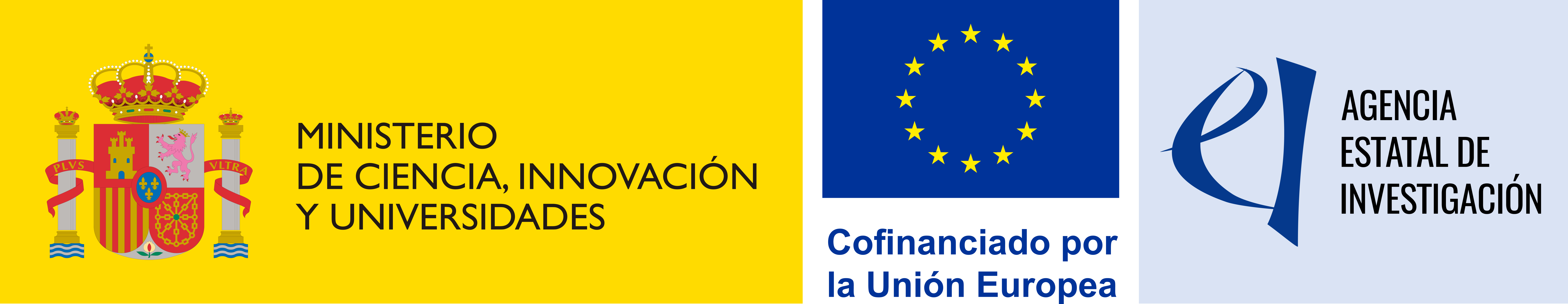}\\[-0.0em] 
\includegraphics[width=0.6\linewidth]{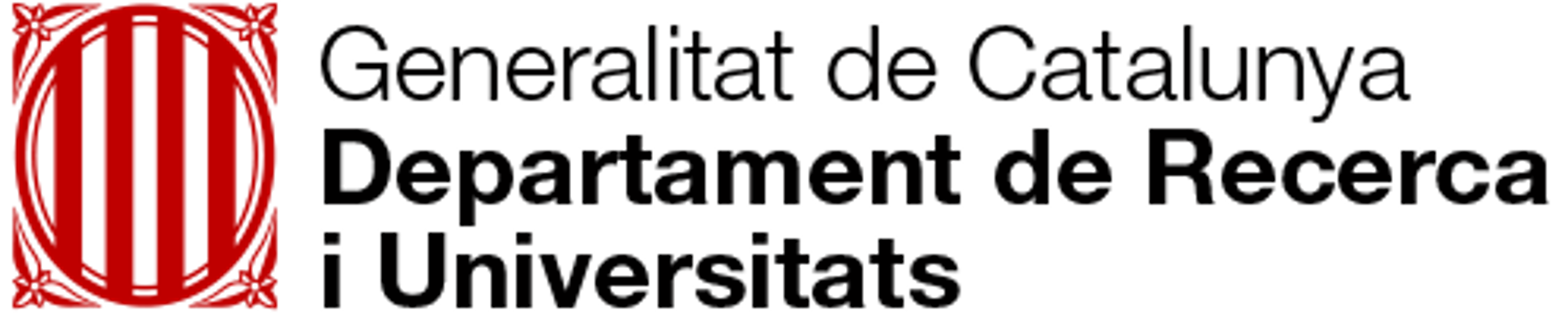}
\end{figure}



\bibliographystyle{unsrt}
\bibliography{biblio}




\end{document}